\documentclass{article}
\usepackage{spconf,amsmath,graphicx}
\usepackage{lipsum}
\usepackage{mathtools}
\usepackage{cuted}
\usepackage{url}
\usepackage{microtype}

\title{Analyzing deep CNN-based utterance embeddings \\ for acoustic model adaptation	}
\name{Joanna Rownicka, Peter Bell, Steve Renals}
\address{The Centre for Speech Technology Research, University of Edinburgh, United Kingdom}

\begin{document}
%
\maketitle
\begin{abstract}

We explore why deep convolutional neural networks (CNNs) with small two-dimensional kernels, primarily used for modeling spatial relations in images, are also effective in speech recognition. We analyze the representations learned by deep CNNs and compare them with deep neural network (DNN) representations and i-vectors, in the context of acoustic model adaptation.
To explore whether interpretable information can be decoded from the learned representations we evaluate their ability to discriminate between speakers, acoustic conditions, noise type, and gender using the Aurora-4 dataset. We extract both whole model embeddings (to capture the information learned across the whole network) and layer-specific embeddings which enable understanding of the flow of information across the network.
We also use learned representations as the additional input for a time-delay neural network (TDNN)  for the Aurora-4 and MGB-3 English datasets. We find that deep CNN embeddings outperform DNN embeddings for acoustic model adaptation and auxiliary features based on deep CNN embeddings result in similar word error rates to i-vectors.

\end{abstract}
\section{Introduction}
\label{sec:intro}

Deep convolutional neural network (CNN) models with small two-dimensional kernels, designed for image recognition~\cite{vgg, imagenet, inception}, have recently been investigated for various speech processing tasks, using speech features organized as a two-dimensional time-frequency matrix.
Earlier works on CNNs  for speech recognition applied convolutional filters solely across the frequency axis either sharing the weights for all frequency bands or with limited weight sharing using separate sets of weights for different frequency bands~\cite{microsoft2013,microsoft2014}. Alternatively, time-delay neural networks (TDNNs) apply convolutions in time in a hierarchical manner and thus are able to exploit variable-length contextual information~\cite{waibel1989, tdnn1}. More recent works have shown good performance of CNN models which convolve in both time and frequency \cite{toth2014}, including very deep networks using stacked small convolutional filters~\cite{ibm,ibm_advances,microsoft2016,microsoft2017,tan2018}. Empirical results have shown that the performance of deep CNNs is comparable to long short term memory (LSTM) recurrent neural networks (RNNs)~\cite{microsoft2017} and compatible to bidirectional LSTM RNNs~\cite{parity}.  The feed-forward nature of CNN models results in lower latency and therefore may be preferable in real-time scenarios~\cite{recent_progress}.

It is hypothesized that localized convolutions across frequency can enable the network to learn speaker-invariant representations by normalizing the spectral variations stemming from  differences in vocal tract lengths, and that convolutions across time can be beneficial in reverberant environments, where temporal artifacts are introduced or to account for speaking rate variation~\cite{time_freq,timefreq}. We investigate whether the use of small two-dimensional (3x3) stacked filters does indeed enable speaker, gender, channel, and noise-invariant representations to be learned. We also compare deep CNN and DNN representations.

Many neural network visualization techniques have been proposed in  computer vision~\cite{feature_vis,interpret_distill,montavon2018}, and can be viewed in three categories: feature visualization, attribution, and dimensionality reduction~\cite{interpret_distill}. Since the input data used for speech recognition is less directly interpretable compared to natural images, we chose to investigate the representations in the activation space in order to understand how the acoustic models represent the data. To interpret the representations learned across the whole network we used dimensionality reduction techniques. We also explored the features learned by different layers in the network to reason about the dynamics of the model learning process. 


We also compared DNN and deep CNN embeddings with i-vectors,  motivated by the fact that the learned representations can be regarded as vectors summarizing the acoustics in the utterance~\cite{summary,xvector}, and hence can be used as an additional input for acoustic model adaptation. In our work, sentence averaging is not in the final component as in~\cite{summary} but information from layers at different positions in the network is combined in order to capture the representations at different levels of abstraction; we also use a deep CNN model in addition to a DNN. Since i-vectors are used for the acoustic model adaptation,  comparisons with i-vectors serve as a guidance on what type of information is desired in the embedding to perform well in attribute-aware training. Finally, we use  conclusions from the analyses of the embeddings and the i-vectors to adapt a TDNN acoustic model.


\section{Utterance embeddings}

Neural network acoustic models primarily learn senone classes using discriminative information from a relatively short acoustic context.  However they can also learn longer-term features, for instance speaker characteristics and additive noise conditions. For this reason we explore learned representations at the utterance level, pooling across multiple frames, which also smooths the representations compared to the frame level.  In addition, we combine the information learned in different layers of the network resulting in a whole model embedding;  we also examine layer-specific embeddings to investigate what information is lost in this combination, as well as to better understand the flow of the information across the network. 



\begin{figure}[ht!]
  \centering
  \centerline{\includegraphics[height=5cm]{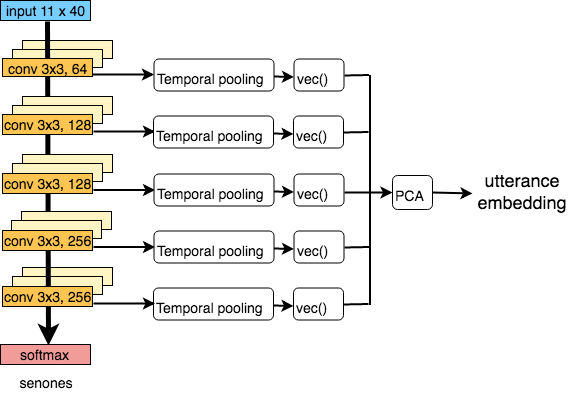}}
  \caption{Deep CNN utterance embedding extraction framework. We take  frame-level activations averaged over utterance for all channels of the last layer of each convolutional block.  These representations are vectorized and concatenates, and PCA is used to reduce the dimensionality.}
\label{fig:vdcnn_embed}
\end{figure}

We extracted the embeddings from trained DNN and deep CNN models, trained using 40-dimension mel filter bank (FBANK) features using $\pm$5 frames of context, with mean and variance normalization. The DNN is a 6 layer network with 2048 nodes in each layer. Following~\cite{my}, the deep CNN consists of 15 convolutional layers, with 3 layers in each convolutional block (using 3x3 kernels), with the number of channels doubling or staying constant for each convolutional block (Fig.~\ref{fig:vdcnn_embed}).  Both models used a ReLU activation function. After training on the Aurora-4 corpus (\url{aurora.hsnr.de}) with multi-condition training, this resulted in word error rates (WERs) of 2.32\%, 5.45\%, 5.38\%, and 15.56\% for test sets A, B, C, and D respectively. For MGB-3 English corpus (\url{www.mgb-challenge.org}), the WER for a deep CNN model for the dev17a test set was 36.7\%.


Fig.~\ref{fig:vdcnn_embed} shows the utterance embedding extraction framework for a deep CNN model. To extract the utterance-level embeddings, we perform a forward pass of the frame-level input features through a trained network.  We average the frame level output of each convolutional block across utterance (temporal pooling) before applying the ReLU function so that their distribution is closer to normal, and hence better suited for further processing. We then vectorize and concatenate each block output, resulting in whole model utterance-level embedding vectors that merge the information across layers. For the deep CNN model this results in a  35k-dimensional vector, and for the DNN a 12k-dimensional vector. To obtain the final utterance representation we reduce the dimensionality to a few hundred using principal components analysis (PCA).

To obtain layer-specific embeddings, we pool the frame-level activations before applying the ReLU function for each test utterance, similar to the whole model embeddings. We explore 5 deep CNN and 6 DNN layers at different positions in the network. We also evaluate the input and output representations pooled in time for each utterance in the speaker, acoustic condition, noise type, and gender recognition task. We hypothesize that input and output representations would be less characteristic of those attributes than the intermediate representations.

\section{Analyzing utterance embeddings}

To investigate the information contained in the utterance embeddings we explore their use for the identification of speaker, acoustic condition, noise type, and gender using Aurora-4.  We use three simple classification functions -- cosine distance, linear discriminant analysis (LDA), and probabilistic LDA (PLDA), evaluating using equal error rate (EER).
We also compare our embeddings with i-vectors~\cite{dehak2011front}, which also model all of the variabilities together via the total variability matrix. 
We extract i-vectors on FBANK features to match the features used in the utterance embedding extraction, as well as on mel-frequency cepstral coefficient (MFCC) features to obtain better quality i-vectors for fair comparison, in both cases using a full-covariance universal background model (UBM) with 2048 components. The i-vector dimensionality matches the dimensionality of the utterance vectors (400). I-vectors are extracted per utterance, similar to our proposed deep embeddings. 


We use Aurora-4 dataset to evaluate the embeddings. The multi-condition training set consists of 7137 utterances coming from 83 speakers and 14 acoustic conditions. We use this set for PLDA and LDA training.  We use standard Aurora-4 test set to create two disjoint sets: evaluation and enrollment. Both sets consist of 2310 utterances chosen randomly from Aurora-4 test set, such that all 8 speakers are present in both sets and the number of utterances per speaker are balanced across both sets. Speakers in the training set do not overlap with the enrollment (enroll) and evaluation (eval) datasets. Acoustic conditions in the training set match the conditions in enroll and eval datasets. The target/nontarget proportion for the trials used for scoring is 50\%, which results in 4620 trials. Each evaluation vector at the utterance level is scored against the enrollment vectors averaged for speakers or acoustic conditions. 


\section{Whole model embeddings}

Whole model embeddings are designed to capture all of the information learned by the NN models by projecting to a common feature space. We evaluate them using speaker recognition and acoustic condition recognition tasks and we compare them to i-vectors -- which are representations commonly used to  characterize utterances, and have been proven to perform well as auxiliary features in speech recognition \cite{saon2013speaker}. These evaluations aim to demonstrate the differences between different utterance-level representations in terms of the amount of interpretable information that they contain.

We use PCA to reduce the dimensionality of the whole model embeddings to 400; the dimensionality of 400 was chosen such that the amount of variance to be explained by all of the selected components is greater than 99.9\%. Table~\ref{tab:components} shows the distribution of the highest variance components by layers for the  DNN and the deep CNN whole model embeddings. The PCA components in the deep CNN embedding are more evenly distributed among layers, whereas the majority of the highest variance PCA components in the DNN embedding come from the last fully-connected layer. This result already suggests that the representation learning mechanism is different between the two models. Further experiments aim to investigate those differences in more detail.

\begin{table}[t]
        \caption{\label{tab:components} {\it Distribution of the highest variance PCA components by layers for 400-dimensional DNN and deep CNN whole model embeddings.}}
        \vspace{2mm}
        \centerline{
          \begin{tabular}{lr|lr}
            \hline
             DNN &  & deep CNN &  \\
            \hline
            \emph{FC0} & 3.25\% & \emph{conv1} & 13.50\% \\ 
            \emph{FC1} & 3.50\% & \emph{conv2} & 13.75\% \\
            \emph{FC2} & 1.50\% & \emph{conv3} & 13.25\% \\
            \emph{FC3} & 0.50\% & \emph{conv4} & 22.75\% \\
            \emph{FC4} & 5.00\% & \emph{conv5} & 36.75\% \\
            \emph{FC5} & 86.25\% &  &  \\
          \end{tabular}
       }
\end{table}

\subsection{Speaker recognition}

\begin{table}[t]
        \caption{\label{tab:spk} {\it EER (\%) for i-vectors and deep embeddings evaluated in the speaker recognition task with different backends. }}
        \vspace{2mm}
        \centerline{
          \begin{tabular}{l|ccc}
            \hline
              & \multicolumn{3}{c}{\emph{EER/\%}} \\
              & cosine & LDA & LDA/PLDA \\
            \hline
            \textsl{i-vector (FBANK)} & 14.76 & 6.93 & 4.55 \\
            \textsl{i-vector (MFCC)} & \textbf{5.71} & 4.20 & 0.74 \\ 
            \hline
            \textsl{DNN embed.} & 35.71 & 6.06 & 2.25 \\
            \textsl{deep CNN embed.} & 22.68 & \textbf{1.65} &  \textbf{0.39} \\
			\hline
          \end{tabular}
       }
\end{table}

Table~\ref{tab:spk} shows the results of applying the  extracted embeddings, as well as i-vectors, to the speaker recognition task. For the LDA and LDA/PLDA backends we use gender independent models. To perform cosine scoring we compute the dot product between the enrollment speaker vectors (utterance vectors averaged per speakers) and evaluation vectors at the utterance level, and we subtract the global mean calculated over the training set from the enrollment and evaluation vectors. In the  LDA/PLDA backend LDA is used to decrease the dimensionality prior to PLDA; this was optimized for each vector type separately (50 for FBANK i-vectors, 70 for MFCC i-vectors, 55 for DNN embeddings, and 70 for deep CNN embeddings). In the LDA backend, the dimensionality was 10 for all vector types. Both i-vectors and our proposed DNN and deep CNN embeddings are computed per utterance and are length normalized.  Increased speaker recognition accuracy corresponds to a reduction in EER.

I-vectors computed on MFCC features give the best performance when evaluated with a cosine similarity backend. Here, all of the dimensions of the vectors being scored are used. This result suggests that ``raw'' i-vectors computed on MFCC features contain much more information about the speakers compared to raw deep embeddings. However, using LDA or PLDA, which make the use of the training speaker labels to learn speaker transforms, brings much more favorable results to deep embeddings, especially deep CNN embeddings -- the lowest obtained EER was for deep CNN embeddings with an LDA/PLDA backend (0.39\%).  MFCC i-vectors consistently result in a lower EER compared to DNN embeddings.  This result suggests that deep CNN embeddings may be highly effective for speaker recognition task, although more experiments using a dataset with a greater number of speakers would be necessary to confirm this.

\subsection{Acoustic condition recognition}

To learn whether deep embeddings contain the information sufficient to differentiate between 14 acoustic conditions (utterances with 7 different types of noise added, recorded with matched or mismatched microphone) we also perform an acoustic condition recognition task. The acoustic conditions in the training set are of the same type as the conditions in enroll and eval datasets, however the noise was added to each utterance at randomly chosen 5-15 dB SNR level for the eval and enroll utterances and 10-20 dB SNR level for the multi-condition training set. This introduces variability to the data and hence the acoustic conditions in the training set are not fully matched to the enroll and eval sets. In this experiment, we average utterance-level vectors across the acoustic conditions to obtain enroll vectors.
Table~\ref{tab:ac} shows the EERs for all four types of embeddings using cosine distance and LDA (with a speaker informed LDA transform).

\begin{table}[t]
        \caption{\label{tab:ac} {\it EER (\%) for i-vectors and deep embeddings evaluated in the acoustic condition recognition task. LDA is informed by both genders speaker labels. }}
        \vspace{2mm}
        \centerline{
          \begin{tabular}{lrrr}
            \hline
             & cosine & LDA (10) & LDA (70) \\
            \hline
            \emph{i-vectors (FBANK, 400)} & 25.97 & 42.51 & 31.60 \\ 
            \emph{i-vectors (MFCC, 400)} & 16.15 & 45.50 & 24.50 \\
            \hline
            \emph{DNN embed. (400)} & \textbf{10.13} & 45.89 & 31.21 \\
            \emph{deep CNN embed. (400)} & 10.95 & 47.88 & 38.87 \\
			\hline
          \end{tabular}
       }
\end{table}

The lowest EER (10.13\%) was achieved using a raw DNN embedding, suggesting that the representations learned by a DNN model contain the most information about the acoustic noise. Deep CNN embeddings also seem to be able to differentiate between acoustic conditions well; however, i-vectors perform more poorly. When we transform the embeddings using the LDA speaker transform, all of the representations lose their ability to differentiate between acoustic conditions to some degree. The LDA experiments were carried out with a dimensionality of 10 (optimal for LDA gender independent scoring of deep CNN embeddings for speaker recognition) and of 70 (optimal for gender independent scoring of MFCC i-vectors).  Reducing the dimensionality of the embeddings with speaker LDA transform results in more speaker-informative and less domain-informative embeddings. We use this observation in Sec.~\ref{sec:adapt} to improve the quality of the extracted embeddings for acoustic model adaptation by forcing the embeddings to be more similar to i-vectors in terms of the encapsulated information.

Comparing the results using the cosine distance backend for speaker and acoustic condition recognition within the same embedding type suggests which attributes of the speech signal are modeled by those representations. Both MFCC and FBANK i-vectors are more speaker specific than noise specific, with MFCC i-vectors being much better for both tasks compared to FBANK i-vectors. However, the results show the opposite for deep embeddings: both DNN and deep CNN embeddings result in a lower EER for acoustic condition recognition.  The biggest difference between DNN and deep CNN representations is in their ability to characterize speakers, with deep CNN embeddings being superior for this task. These results suggest that the deep CNN acoustic model might perform better in the ASR task compared to the DNN model (9.55\% WER compared to 12.55\% WER) because of the deep CNN model's ability to learn more speaker-aware intermediate representations. A possible reason for learning more speaker-aware representations by deep CNN models compared to DNNs is weight sharing in time and frequency which can be contributing to capturing more speaker characteristic features. 

\section{Layer-specific embeddings}

Layer-specific embeddings capture the information contained in specific layers in the network. By looking at the discriminative power of these representations we aim to learn more about the learning process of a deep CNN model, as well as about a DNN model for comparison. Besides the comparison between different models, it is also interesting to compare the layer-specific embeddings within the same model to see how the information about different attributes is propagated in the networks.

We choose to keep the dimensionality of layer-specific embeddings constant (80), however we also examined the number of components needed in the layer-specific PCA representation in order to retain 99\% of its descriptive power. For the DNN model, the proportion of components required in subsequent fully-connected layers is 1\%, 3\%, 8\%, 14\%, 21\%, 31\%. This confirms that the representations in the upper layers are richer, thus more components are needed to represent the same amount of variability. This observation is also confirmed for the deep CNN model, with 1\%, 1\%, 3\%, 6\%, 14\% of components needed to retain the variability at a constant 99\%.

The results for DNN layers are presented in table~\ref{tab:layerDNN}, and the representations for the last convolutional layer in each of the 5 convolutional blocks of the deep CNN are in table~\ref{tab:layerCNN}. The results for the embeddings of variable number of components (but with the constant variance explained) followed the same pattern as fixed size embeddings. We provide the EER score for input and output representations as well, which were obtained similarly to the utterance-level embeddings by averaging frame-level representations for each utterance.

The layer-specific embeddings confirm the findings from the previous section. Deep CNN embeddings perform much better than DNN embeddings in speaker characterization and similarly in acoustic condition recognition. DNN embeddings have a lower EER for recognizing the noise type, but are worse in differentiating between genders. DNN internal representations are more aware of the background acoustic noise (acoustic condition and noise recognition scores), and deep CNN representations are more aware of the speaker related characteristics (speaker and gender recognition scores). The deep CNN model learns about speaker and gender characteristics much faster than the DNN model. After only the first convolutional block (3 layers) the EER drops from 48.35\% at the input to 27.53\% for speaker recognition. For the DNN model, the EER after 3 layers drops to 34.68\%. Using small convolutional kernels in time and frequency and sharing the weights across the whole input feature map(s) enables the network to learn more speaker-aware representations compared to densely connected DNN models without weight sharing. Speaker-aware representations are also learned faster, and the ability of the layer-specific embeddings to characterize speakers degrades after the third convolutional block. In order to perform well in the ASR task, the network has to learn about the speakers first (first three convolutional blocks), to then solve the senone classification task, which results in the internal representations less representative of a speaker (blocks 4 and 5). The deep CNN network is therefore performing speaker normalization. The same applies to gender normalization, which is performed by the DNN model as well. This result shows that gender normalization is an easier task than speaker normalization, however, deep CNN model is more effective than the DNN model in gender normalization. Deep CNN model is not performing noise normalization. 

At the final convolutional block of the deep CNN model (conv5), the network has less knowledge about the speakers than in the middle layer (conv3). However, the majority of the highest variance components in the whole model embedding are in the last block. We hypothesize that this variance is due to the phonetic information in the utterances which would explain why deep CNN perform better in the speech recognition task than DNN models. For the DNN model, the majority of the highest variance components are also in the last layer, but the representation at the last FC layer is not a speaker-invariant representation -- the ability to differentiate between speakers is the highest at the last layer, which might explain worse performance compared to deep CNN model.


\begin{table}[t]
        \caption{\label{tab:layerDNN} {\it EER (\%) with cosine distance scoring for layer-specific embeddings in a DNN model. The dimensionality for each layer is specified in the parenthesis.}}
        \vspace{2mm}
        \centerline{
          \begin{tabular}{lrrrr}
            \hline
             & spk & ac. cond. & noise & gender \\
            \hline
            \emph{input (80)} & 48.35 & 20.95 & 19.35 & 49.31 \\ 
            \emph{FC0 (80)} & 47.32 & 11.65 & 9.09 & 49.22 \\ 
            \emph{FC1 (80)} & 44.16 & 11.95 & 10.00 & 46.02 \\ 
            \emph{FC2 (80)} & 34.68 & 11.00 & 8.14 & 33.98 \\ 
            \emph{FC3 (80)} & 31.21 & 10.69 & 7.75 & \textbf{31.43} \\ 
            \emph{FC4 (80)} & 30.65 & 10.30 & 7.92 & 35.93\\ 
            \emph{FC5 (80)} & \textbf{27.92} & \textbf{10.17} & \textbf{6.71} & 37.88 \\ 
            \emph{output (80)} & 40.26 & 22.42 & 21.08 & 44.29 \\ 
			\hline
            \emph{whole model (400)} & 35.71 & 10.13 & 7.32 & 38.10 \\ 
            \hline
          \end{tabular}
       }
\end{table}


\begin{table}[t]
        \caption{\label{tab:layerCNN} {\it EER (\%) with cosine distance scoring for layer-specific embeddings in a CNN model. The dimensionality for each layer is specified in the parenthesis.}}
        \vspace{2mm}
        \centerline{
          \begin{tabular}{lrrrr}
            \hline
             & spk & ac. cond. & noise & gender \\
            \hline
            \emph{input (80)} & 48.35 & 20.95 & 19.35 & 49.31 \\ 
            \emph{conv1 (80)} & 27.53 & 11.47 & 9.61 & 27.40 \\ 
            \emph{conv2 (80)} & 23.38 & 11.17 & 10.39 & 26.75 \\ 
            \emph{conv3 (80)} & \textbf{20.65} & \textbf{10.87} & 9.48 & \textbf{20.22} \\ 
            \emph{conv4 (80)} & 25.54 & 11.56 & 9.57 & 26.71 \\ 
            \emph{conv5 (80)} & 34.20 & 11.77 & \textbf{8.57} & 42.47 \\ 
            \emph{output (80)} & 40.22 & 24.42 & 25.06 & 46.10 \\ 
			\hline
            \emph{whole model (400)} & 22.68 & 10.95 & 9.18 & 24.50 \\ 
            \hline
          \end{tabular}
       }
\end{table}

\section{Utterance embeddings for acoustic model adaptation}
\label{sec:adapt}

The experiments in this section examine the use of  whole model utterance embeddings for the task of acoustic model adaptation, using a TDNN acoustic model. The analyses in the previous sections of the paper enabled us to better understand the type of information contained in the deep embeddings, and by comparing the embeddings to i-vectors we gained an intuition on what a representation suitable for acoustic model adaptation should represent. We train the TDNN baselines for Aurora-4 and for  MGB-3, and for each examine different types of auxiliary features to inform the network about the attributes of the utterances.

The TDNN baseline for Aurora-4 is trained on raw 13-dimensional MFCC features, without mean and variance normalisation. It comprises 5 TDNN layers each with 650 units. The model has a left context of 13 and right context of 7. All Aurora-4 models use the alignments generated by a triphone GMM model. The WERs for Aurora-4 TDNN baseline as well as for models using different embeddings to inform the network about the utterance attributes are presented in table~\ref{tab:adaptAurora}. 

In noisy conditions (test sets B and D), i-vectors computed on MFCC features perform  best. They also were the most speaker characteristic embeddings if no additional LDA or PLDA transforms were used. Thus, using the most speaker-aware auxiliary features appears to be the best choice for adaptation in noisy conditions. Further results confirm this finding. Providing a representation which is more noise-aware but less speaker-aware (raw DNN and deep CNN embeddings) degrades the ability of the acoustic model to recognize tied-state triphones. We can recover some of the model's performance by applying an LDA transform to the embeddings. An LDA transform projects the embeddings into a lower-dimensional space with good between-speaker separability, hence the DNN and deep CNN embeddings transformed with LDA are more speaker-aware (as seen in table~\ref{tab:spk}) and less noise-aware (as seen in table~\ref{tab:ac}). They are then better suited for acoustic model adaptation task (table~\ref{tab:adaptAurora}). For clean test sets (A and C) LDA transformed NN embeddings outperform i-vectors. Also, their dimensionality is much smaller (10 dimensions). 

We also use the embeddings as auxiliary features for the MGB-3 task. The TDNN baseline for MGB-3 is a 5 layer TDNN model with 1280 units in each layer. The input features are 40-dimension high resolution MFCCs, appended with 3 pitch features. The model's left and right context is 11 and 9 frames respectively. All MGB-3 TDNN models use the same alignments, generated from a sequence discriminative trained TDNN model with i-vector adaptation. The results for the MGB-3 dataset are in table~\ref{tab:adaptMGB}. 100-dimension i-vectors extracted on FBANK features were the representations providing the most gains over the baseline TDNN model. Similar to the results for Aurora-4, using raw DNN or deep CNN embeddings as auxiliary features resulted in a performance degradation (WER worse than unadapted baseline or close to it). However, we found that using the model trained for 2 epochs instead of 4, resulted in a lower WER. 
Further analysis was guided by the results from Aurora-4 adaptation experiments. We used the LDA transform trained on MGB-3 training set labels corresponding to colors of the captions to improve the speaker-awareness of the embeddings. This strategy proved to be effective. The WER for all segments for the LDA transformed deep CNN embeddings match the WER for 400-dimensional MFCC i-vectors but with much lower dimensionality (70).

\begin{table}[tb]
        \caption{\label{tab:adaptAurora} {\it WER (\%) for Aurora-4 test sets.}} 
        \vspace{2mm}
        \centerline{
          \begin{tabular}{lrrrr}
            \hline
            \textbf{Model} & \textbf{A} & \textbf{B} & \textbf{C} & \textbf{D} \\
            \hline
            \emph{TDNN baseline} & 3.64 & 7.69 & 8.96 & 19.45 \\ \hline  
            \emph{+ FBANK i-vectors (400)} & 3.77 & 7.27 & 8.22 & 18.76 \\ 
            \emph{+ MFCC i-vectors (400)} & 3.71 & \textbf{6.99} & 7.86 & \textbf{17.56} \\ \hline 
            \emph{+ DNN embedding (400)} & 4.38 & 8.22 & 12.15 & 20.12 \\  
            \emph{+ deep CNN embedding (400)} & 4.54 & 8.02 & 9.37 & 20.05 \\ \hline 
            \emph{+ LDA (10) DNN embed.} & \textbf{3.58} & 7.53 & 8.76 & 18.79 \\ 
            \emph{+ LDA (10) deep CNN embed.} & 3.65 & 7.37 & \textbf{7.77} & 17.99 \\ 
            \hline
          \end{tabular}
       }
 \end{table}

\begin{table}[t]
        \caption{\label{tab:adaptMGB} {\it WER (\%) for MGB-3 English dev17a test set. 2e means training for 2 epochs.}}
        \vspace{2mm}
        \centerline{
          \begin{tabular}{lr}
            \hline
            \textbf{Model} & \textbf{WER} \\
            \hline
            \emph{TDNN baseline} &  29.4 \\ 
            \hline
            \emph{+ FBANK i-vectors (400)} & 27.9 \\ 
            \emph{+ FBANK i-vectors (100)} & \textbf{27.2} \\ 
            \emph{+ MFCC i-vectors (400)} & 27.7 \\ 
            \emph{+ MFCC i-vectors (100)} & 27.5 \\ 
            \hline
            \emph{+ DNN embed. (400)} & 29.8 \\ 
            \emph{+ DNN embed. (400, 2e)} & 28.6 \\ 
            \hline
            \emph{+ deep CNN embed. (400)} & 29.3 \\ 
            \emph{+ deep CNN embed. (400, 2e)} & 28.6 \\ 
            \hline
            \emph{+ LDA (10) DNN embed.} & 28.8 \\ 
            \emph{+ LDA (10) deep CNN embed.} & 28.7 \\ 
            \emph{+ LDA (70) deep CNN embed.} & \textbf{27.7} \\ 
            \emph{+ LDA (70) deep CNN embed. (2e)} & 28.4 \\ 
            \hline
          \end{tabular}
       }
 \end{table}

\section{Conclusions and discussion}

In this work we analyzed the representations learned by deep CNN models and compared them to DNN representations to better understand the differences in the learning process between those two models. We find that deep CNN models are able to learn more speaker-, gender-, and channel-invariant representations than DNN models. This means that the better performance of the CNN models stems from a better problem representation and deep CNNs will potentially perform better than DNNs when applied to new speakers and in  mismatched channel conditions. On the other hand, it seems that a limitation of both types of models lies in learning noise-invariant representations. Addressing this issue may contribute to further improvements for ASR in noisy conditions. Looking at the representative power at different layers in the network as shown in this paper can help to determine the appropriate model for the task at hand. 

We also compared the extracted embeddings with i-vectors in order to better understand the difference in the type of information captured by both types of acoustic summaries of an utterance. Using the embeddings in their raw form improved the ASR results over the non-adapted baseline model. It is important to note that this method does not involve any speaker information and it also does not require an additional i-vector extraction framework. Because the labels of the extractor match the labels used in the main acoustic model, joint training of the extractor and the main model should be possible. 

We further used the analyses of the embeddings and their comparison with i-vectors as a guide to construct a more informative auxiliary feature vector for the acoustic model adaptation task. At this point, speaker labels are required, and with a speaker-informed LDA transform we were able to further improve the performance of the embeddings for the acoustic model adaptation. For Aurora-4 they outperform i-vectors for test sets without the additive noise. For MGB-3 English, they don't outperform the best performing i-vectors, but do match the performance of the 400-dimensional i-vectors extracted on top of the MFCC features. One possible explanation is that the LDA transform for MGB-3 was not good enough. We plan to experiment with more reliable speaker labels for LDA transform extraction to test this hypothesis.

Embeddings extracted in a way presented in this paper can be regarded as a generic framework which is able to produce the acoustic summary vectors for sequential data. There are therefore other possible use cases for those embeddings, other than the acoustic model adaptation, for instance the selection of the augmented training data because of the embeddings ability to well differentiate different acoustic conditions, or as a similarity measure for fMLLR initialization. 

The other possible future direction is the improvement of the design of the embeddings for the acoustic model adaptation. In this work, we extracted one embedding per utterance to be able to analyze the attributes at the utterance level. We hypothesize that introducing more variability to the embeddings at training time will be beneficial for the model adaptation. We plan to investigate the embeddings extracted every couple of frames for training and utterance level embeddings at test time for a more optimal setting. We also plan to experiment with the attention mechanism instead of the average pooling operation, and with the use of different types of models (e.g. LSTM) as the extractors. 

\bigskip
\noindent
\textbf{Acknowledgement:}
This work was supported by a PhD studentship from the DataLab Innovation Centre, Ericsson Media Services, and Quorate Technology.

\clearpage
\bibliographystyle{IEEEbib}
\bibliography{refs}

\end{document}